\documentclass{article}
\pdfoutput=1

\usepackage[final]{neurips_2018}

\usepackage[utf8]{inputenc} 
\usepackage[T1]{fontenc}  
\usepackage[draft]{hyperref}     
\usepackage{url}      
\usepackage{booktabs}     
\usepackage{amsfonts}     
\usepackage{nicefrac}     
\usepackage{microtype}    

\usepackage{amsmath,amssymb,amsthm}
\usepackage[usenames,dvipsnames,svgnames,table]{xcolor}
\usepackage{graphicx}
\usepackage{wrapfig}
\usepackage{stmaryrd}

\usepackage{tikz}
\usetikzlibrary{trees}
\usepackage{forest}

\usepackage{subcaption}
\usepackage{listings} 

\usepackage{appendix}
\usepackage[capitalise]{cleveref} 
\crefname{listing}{Algorithm}{Algorithm}
\Crefname{listing}{Algorithms}{Algorithms}

\theoremstyle{definition}
\newtheorem{theorem}{Theorem}
\newtheorem{lemma}[theorem]{Lemma}

\newcommand{\indicator}[1]{\llbracket #1 \rrbracket}

\definecolor{darkred}{rgb}{.7,0,0}
\definecolor{darkgreen}{rgb}{0,.5,0}
\definecolor{darkblue}{rgb}{0,0,.8}
\definecolor{darkcyan}{rgb}{0,0.6,.6}
\definecolor{darkorange}{rgb}{.8,.4,0}
\definecolor{gray}{rgb}{.4,.4,.4}

\newcommand{\Naturals}{\mathbb{N}_1}
\newcommand{\Nonnegints}{\mathbb{N}_0}
\newcommand{\Reals}{\mathbb{R}}

\newcommand{\ie}{\emph{i.e.}, }
\newcommand{\eg}{\emph{e.g.}, }

\DeclareMathOperator*{\expect}{\mathbb{E}}
\DeclareMathOperator*{\argmax}{\text{argmax}}
\DeclareMathOperator*{\argmin}{\text{argmin}}

\newcommand{\eps}{\varepsilon}
\newcommand{\half}{\tfrac{1}{2}}

\author{
Laurent Orseau \\ DeepMind, London, UK \\ \texttt{lorseau@google.com}
\And
Levi H. S. Lelis\thanks{This work was carried out while L. H. S. Lelis was at the University of Alberta, Canada.} \\ 
Universidade Federal de Viçosa, Brazil \\
\texttt{levi.lelis@ufv.br} 
\AND
Tor Lattimore \\ DeepMind, London, UK \\ \texttt{lattimore@google.com}
\And
Th\'eophane Weber \\ DeepMind, London, UK \\ \texttt{theophane@google.com}
}

\newcommand{\prog}{\rho}

\newcommand{\tree}{\mathcal{T}}
\newcommand{\node}{n}
\newcommand{\children}{\mathcal{C}}
\newcommand{\rootnode}{\node_0}
\newcommand{\nodeset}{\mathcal{N}}
\newcommand{\leaves}{\mathcal{L}}
\newcommand{\targetset}{\nodeset^g}
\newcommand{\targetnode}{\node^g}
\newcommand{\state}{s}
\newcommand{\stateset}{\mathcal{S}}
\newcommand{\goalset}{\stateset^g} 
\newcommand{\act}{a}
\newcommand{\actionset}{\mathcal{A}}
\newcommand{\pol}{\pi}
\newcommand{\depth}{d}  
\newcommand{\depthz}{d_0}  

\newcommand{\rootstate}{\state_0}
\newcommand{\trans}{T}   

\newcommand{\nodexp}{N}  
\newcommand{\levints}{\text{LevinTS}}
\newcommand{\lubyts}{\text{LubyTS}}
\newcommand{\multits}{\text{multiTS}}
\newcommand{\magicseq}{\text{A6519}}

\newcommand{\prob}{p}
\newcommand{\cumu}{q}
\newcommand{\runt}{t} 
\newcommand{\uprunt}{\hat{\runt}} 

\newcommand{\depthbound}{{d_{\text{max}}}} 

\newcommand{\fringe}{\mathcal{F}}
\newcommand{\visited}{\mathcal{V}}
\newcommand{\cost}{\text{cost}}
\newcommand{\TS}{\text{TS}}

\ifdefined\suppmat
\title{Single-Agent Policy Tree Search With Guarantees: \\ Supplementary Material}
\else
\title{Single-Agent Policy Tree Search With Guarantees}
\fi

\begin{document}

\maketitle

\ifdefined\suppmat\newpage\fi

\begin{abstract}
We introduce two novel tree search algorithms that use a policy to guide search. The first algorithm is a best-first \emph{enumeration} that uses a cost function that allows us to prove an \emph{upper bound} on the number of nodes to be expanded before reaching a goal state. We show that this best-first algorithm is particularly well suited for ``needle-in-a-haystack'' problems. The second algorithm is based on \emph{sampling} and we prove an \emph{upper bound on the expected} number of nodes it expands before reaching a \emph{set} of goal states. We show that this algorithm is better suited for problems where many paths lead to a goal. We validate these tree search algorithms on 1,000 computer-generated levels of Sokoban, where the policy used to guide the search comes from a neural network trained using A3C. Our results show that the policy tree search algorithms we introduce are competitive with a state-of-the-art domain-independent planner that uses heuristic search. 
\end{abstract}

\section{Introduction}

Monte-Carlo tree search (MCTS) algorithms~\citep{coulom2007efficient,browne2012survey} have been recently applied with great success to several problems such as Go, Chess, and Shogi~\citep{silver2016mastering,silver2017mastering}.
Such algorithms are well adapted to stochastic and adversarial domains, due to their sampling nature and the convergence guarantee to min-max values. 
However, the sampling procedure used in MCTS algorithms is not well-suited for other kinds of problems~\citep{nakhost2013random}, such as deterministic single-agent problems where the objective is to find any solution at all.
In particular, if the reward is very sparse---for example the agent is rewarded only at the end of the task---MCTS algorithms revert to uniform search.
In practice such algorithms can be guided by a heuristic but, to the best of our knowledge, no bound is known that depends on the quality of the heuristic.
For such cases one may use instead other traditional search approaches such as A*~\citep{hart1968aFormalBasis} and Greedy Best-First Search (GBFS)~\citep{Doran1966}, which are guided by a heuristic cost function.

In this paper we tackle single-agent problems from the perspective of policy-guided search. 
One may view policy-guided search as a special kind of heuristic search in which a policy, instead of a heuristic function, is provided as input to the search algorithm. 
As a policy is a probability distribution over sequences of actions, this allows us to provide theoretical guarantees that cannot be offered by \emph{value} (\eg reward-based) functions:
we can bound the number of node expansions---roughly speaking, the search time---depending on the probability of the sequences of actions that reach the goal.
We propose two different algorithms with different strengths and weaknesses. 
The first algorithm, called \levints{}, is based on Levin search~\citep{levin1973search}
and we derive a strict upper bound on the number of nodes to search before finding the least-cost solution. 
The second algorithm, called \lubyts{}, is based on the scheduling of \citet{luby1993speedup} for randomized algorithms and we prove an upper bound on the expected number of nodes to search before reaching any solution while taking advantage of the potential multiplicity of the solutions. 
\levints{} and \lubyts{} are the first policy tree search algorithms with such guarantees. Empirical results on the PSPACE-hard domain of Sokoban~\citep{Culberson1999} show that \lubyts{} and in particular \levints{} guided by a policy learned with A3C~\citep{mni2016asynchronous} are competitive with a state-of-the-art planner that uses GBFS~\citep{hoffmannN01}. 
Although we focus on deterministic environments, \levints{} and \lubyts{} can be extended to stochastic environments with a known model.

\levints{} and \lubyts{} bring important research areas closer together. Namely, areas that traditionally rely on heuristic-guided tree search with guarantees such as classical planning and areas devoted to learn control policies such as reinforcement learning. We expect future works to explore closer relations of these areas, such as the use of \levints{} and \lubyts{} as part of classical planning systems.

\section{Notation and background}

We write $\Naturals=\{1, 2, \ldots\}$.
Let $\stateset$ be a (possibly uncountable) set of states,
and let $\actionset$ be a finite set of actions.
The environment starts in an initial state $\rootstate\in\stateset$.
During an \emph{interaction step} (or just step)
the environment in state $\state\in\stateset$
receives an action $\act\in\actionset$ from the searcher 
and transitions deterministically 
according to a transition function $\trans:\stateset\times\actionset\to\stateset$
to the state $\state' = \trans(\state, \act)$.
The state of the environment after a sequence of actions $\act_{1:t}$ is 
written $\trans(\act_{1:t})$
which is a shorthand for the recursive application of the transition function $\trans$
from the initial state $\rootstate$ to each action of $\act_{1:t}$,
where $\act_{1:t}$ is the sequence of actions $\act_1, \act_2, \ldots \act_t$.
Let $\goalset\subseteq\stateset$ be a set of goal states.
When the environment transitions to one of the goal states, the problem is solved and the interaction stops.
We consider \emph{tree} search algorithms and 
define the set of nodes in the tree as the set of sequences of actions $\nodeset := \actionset^*\cup\actionset^{\infty}$.
The root node $\rootnode$ is the empty sequence of actions.
Hence a sequence of actions $\act_{1:t}$ of length $t$
is uniquely identified by a node $\node\in\nodeset$
and we define $\depthz(\node)=\depthz(\act_{1:t}) := t$ (the usual depth $\depth(\node)$ of the node is recovered with $\depth(\node) = \depthz(\node) - 1$).
Several sequences of actions (hence several nodes) can lead to the same state of the environment, and we write $\nodeset(\state):=\{\node\in\nodeset:\trans(\node) = \state\}$ for the set of nodes with the same state.
We define the set of children $\children(\node)$ of a node $\node\in\nodeset$
as $\children(\node) := \{\node\act|\act\in\actionset\}$, where $\node\act$ denotes the sequence of actions $\node$ followed by the action $\act$.
We define the \emph{target set} $\targetset\subseteq\nodeset$ as the set of nodes such that the corresponding states are goal states:
$\targetset := \{\node : \trans(\node) \in\goalset \}$.
The searcher does not know the target set in advance and only recognizes a goal state when the environment transitions to one.
If $\node_1 = \act_{1:t}$ and $\node_2 = \act_{1:t}\act_{t+1:k}$ with $k> t$ then we say that
$\act_{1:t}$ is a prefix of $\act_{1:t}\act_{t+1:k}$ and that $\node_1$ is an ancestor of $\node_2$ (and $\node_2$ is a descendant of $\node_1$).

A \emph{search tree} $\tree\in\nodeset^*$ is a set of sequences of actions (nodes)
such that (i) for all nodes $\node\in\tree$, $\tree$ also contains
all the ancestors of $\node$ and (ii) if $\node\in\tree\cap\targetset$, then 
the tree contains no descendant of $\node$.
The leaves $\leaves(\tree)$ of the tree $\tree$ are the set of nodes $\node\in\tree$ such that 
$\tree$ contains no descendant of $\node$.
A \emph{policy} assigns
probabilities to sequences of actions
under the constraint that $\pol(\rootnode)=1$
and $\forall\node\in\nodeset, \pol(\node) = \sum_{\node'\in\children(\node)}\pol(\node')$.
If $\node'$ is a descendant of $\node$, we define the conditional probability $\pol(\node'|\node) := \pol(\node') / \pol(\node)$.
The policy is assumed to be provided as input to the search algorithm.

Let \TS{} be a generic tree search algorithm defined as follows.
At any \emph{expansion step} $k\geq 1$, 
let $\visited_k$ be the set of nodes that have been expanded (visited) before (excluding) step $k$,
and let the fringe set $\fringe_k := \bigcup_{\node\in\visited_k}\children(\node) \setminus \visited_k$ be the set of not-yet-expanded children of expanded nodes,
with $\visited_1 := \emptyset$ and $\fringe_1 := \{\rootnode\}$.
At iteration $k$, the search algorithm \TS{} chooses a node
$\node_k\in\fringe_k$ for \emph{expansion}:
if $\node_k\in\targetset$, then the algorithm terminates with success.
Otherwise, $\visited_{k+1}:=\visited_k \cup \{\node_k\}$ and the iteration $k+1$ starts.
At any expansion step, the set of expanded nodes is a search tree.
Let $\node_k$ be the node expanded by \TS{} at step $k$.
Then we define the search time $\nodexp(\TS, \targetset) := \min_{k>0} \{k:\node_k \in\targetset\}$
as the number of node expansions before reaching any node of the target set $\targetset$.

A policy is \emph{Markovian} if the probability of an action depends only on the current state of the environment,
that is,
for all $\node_1$ and $\node_2$
with $\trans(\node_1) = \trans(\node_2),
\forall \act\in\actionset:\pol(\act|\node_1) = \pol(\act|\node_2)$.
%
In this paper we consider both Markovian and non-Markovian policies.
For some  function $\cost:\nodeset\to\Reals$ over nodes, we define the cost of a state $\state$ as 
$\cost(\state) := \min_{\node\in\nodeset(\state)}\cost(\node)$.
Then we say that a tree search algorithm with a cost function $\cost(\node)$ expands states in \emph{best-first order} if for all states $\state_1$ and $\state_2$, if $\cost(\state_1) < \cost(\state_2)$,
then $\state_1$ is visited before $\state_2$.
We say that a state is expanded at its \emph{lowest cost} if for all states $\state$, the first node $\node\in\nodeset(\state)$ to be expanded has cost $\cost(\node) = \cost(\state)$.

\section{Levin tree search: policy-guided enumeration}

First, we show that merely expanding nodes by decreasing order of their probabilities can fail to reach a goal state of non-zero probability.
\begin{theorem}
The version of \TS{} that chooses at iteration $k$ the node
$\node_k := \argmax_{\node\in\fringe_k} \pol(\node)$
may never expand any node of the target set $\targetset$, 
even if $\forall\node\in\targetset, \pol(\node)>0$.
\end{theorem}

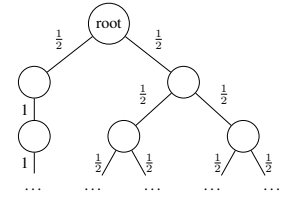
\begin{figure}
\begin{minipage}[t]{0.55\textwidth}
\begin{lstlisting}[numbers=left,numberstyle=\tiny,label=lst:levints,caption={Levin tree search.}]
def LevinTS()
  $\visited := \emptyset$
  $\fringe := \{\rootnode\}$
  while $\fringe \neq \emptyset$
    $\node := \argmin_{\node\in\fringe} \frac{\depthz(\node)}{\pol(\node)}$
    $\fringe := \fringe \setminus\{\node\}$
    $\state := \trans(\node)$
    if $\state\in\goalset$
      return true
    if is_Markov($\pol$)  (* \label{lst:levints:cutstart} *)
      if $\exists \node'\in\visited : (\trans(\node')=\state) \land (\pol(\node') \geq  \pol(\node))$
        # $\state$ has already been visited with
        # a higher probability: State cut
        continue
      $\visited := \visited \cup \{\node'\}$  (* \label{lst:levints:cutstop} *)
    $\fringe := \fringe \cup \children(\node)$
  return false
\end{lstlisting}
\end{minipage}%
\hspace*{0.05\textwidth} 
\begin{minipage}[t]{0.4\textwidth}
\begin{lstlisting}[label=lst:sampling,caption={Sampling and execution of a single trajectory.}]
def sample_traj(depth)
  $\node := \rootnode$
  for d := 0 to depth
    if $\trans(\node) \in\goalset$
      return true
    $\act \sim \pol(.|\node)$
    $\node := \node\act$
  return false
\end{lstlisting}
\begin{minipage}[b]{\textwidth}
\centering
\scalebox{0.6}{
\begin{forest}
for tree={
 grow'=south,
 l=1.2cm,
 s sep=0.6cm,
 minimum width=2em,
 draw,circle,
}
[root,
 [,edge label={node[midway,above right]{$\frac{1}{2}$}}
     [,edge label={node[midway,above right]{$\frac{1}{2}$}}
         [\ldots, draw=none, edge label={node[midway,right]{$\frac{1}{2}$}}]
         [\ldots, draw=none, edge label={node[midway,left]{$\frac{1}{2}$}}]
     ]
     [,edge label={node[midway,above left]{$\frac{1}{2}$}}
         [\ldots, draw=none, edge label={node[midway,right]{$\frac{1}{2}$}}]
         [\ldots, draw=none, edge label={node[midway,left]{$\frac{1}{2}$}}]
     ]
 ]
 [,edge label={node[midway,above left]{$\frac{1}{2}$}},
     [,edge label={node[midway,left]{1}},
         [\ldots,draw=none,edge label={node[midway,left]{1}}]]]]
\end{forest}
}
\caption{A `chain-and-bin' tree.
}
\label{fig:chainandbin}
\end{minipage}
\end{minipage}
\end{figure}

\begin{proof}
Consider the tree in \cref{fig:chainandbin}.
Under the left child of the root is an infinite `chain' in which each node has probability $1/2$.
Under the right child of the root is an infinite binary tree in which each node has two children, each of conditional probability $1/2$, and thus each node has probability $2^{-d}$.
Before testing a node of depth at least 2 in the right-hand-side binary tree (with probability at most $1/4$),
the search expands infinitely many nodes of probability $1/2$.
Defining the target set as any set of nodes with individual probability at most $1/4$ proves the claim.
\end{proof}

To solve this problem, we draw inspiration from Levin search~\citep{levin1973search,trakhtenbrot1984survey},
which (in a different domain) penalizes the probability with computation time.
Here, we take computation time to mean the depth of a node.
The new Levin tree search (\levints) algorithm is a version of \TS{} in which nodes are expanded in order of increasing costs $\depthz(\node) / \pol(\node)$ (see \cref{lst:levints}). 

\levints\ also performs \emph{state cuts} (see Lines~\ref{lst:levints:cutstart}--\ref{lst:levints:cutstop} of \cref{lst:levints}). That is, \levints\ does not expand node $\node$ representing state $\state$ if (i) the policy $\pol$ is Markovian, (ii) it has already expanded another node $\node'$ that also represents $\state$, and (iii) $\pol(\node') \ge \pol(\node)$. By performing state cuts only if these three conditions are met, we can show that \levints{} expands states in best-first order.

\begin{theorem}\label{thm:bestfirstorder}
\levints{} expands states in best-first order and at their lowest cost first.
\end{theorem}
\begin{proof}


Let us first consider the case where the policy is non-Markovian.
Then, \levints{} does not perform state cuts (see Line~\ref{lst:levints:cutstart} of \cref{lst:levints}). 
Let $\node_1$ and $\node_2$ be two arbitrary different nodes (sequences of actions),
with $\cost(\node_1) < \cost(\node_2)$.
Let $\node_{12}$ be the closest common ancestor of $\node_1$ and $\node_2$; it must exist since at least the root is one of their common ancestors.
Then all nodes on the path
from $\node_{12}$ to $\node_1$ have cost less than $\cost(\node_1)$ and thus than $\cost(\node_2)$,
due to the monotonicity of $\depthz$ and $\pol$ and thus of $\cost$,
which implies by recursion from $\node_{12}$ that all these nodes and thus also $\node_1$ are expanded before $\node_2$.
Hence, if $\trans(\node_1)=\trans(\node_2)$, this proves that all states are visited first at their lowest cost.
Furthermore, if $\trans(\node_1)\neq\trans(\node_2)$, this proves that states of lower cost are visited first.

Now, if the policy is Markovian, then we need to show that state cuts do not prevent best-first order and lowest cost. 
Let $\node_1$ and $\node_2$ be two nodes representing the same state $s$, 
where $\node_1$ is expanded before $\node_2$. 
Assume that no cut has been performed before $\node_2$ is expanded.
First, since no cuts were performed, we have from the non-Markovian case that 
$\frac{\depthz(\node_1)}{\pol(\node_1)}\leq \frac{\depthz(\node_2)}{\pol(\node_2)}$.
Secondly, consider a sequence of actions $\act_{1:k}$ taken after state $\state$,
and let $\node_{1k}=\node_1\act_{1:k}$ be the node reached after taking $\act_{1:k}$ starting from $\node_1$
and similarly for $\node_{2k}$.
Since the environment is deterministic,
this sequence leads to the same state $\state_k$, whether starting from  $\node_1$ or from $\node_2$.
Since the policy is Markovian, 
$\pol(\node_{1k}|\node_1) = \pol(\node_{2k}|\node_2)$.
Then from the condition (iii) of state cuts,
\begin{align*}
\text{if } \pol(\node_1) \geq \pol(\node_2),\quad
\frac{\depthz(\node_{1k})}{\pol(\node_{1k})} 
 &= \frac{\depthz(\node_1)}{\pol(\node_1)}\frac{1}{\pol(\node_{1k}|\node_1)} + \frac{k}{\pol(\node_1)\pol(\node_{1k}|\node_1)} \\
&\leq \frac{\depthz(\node_2)}{\pol(\node_2)}\frac{1}{\pol(\node_{1k}|\node_1)} + \frac{k}{\pol(\node_2)\pol(\node_{1k}|\node_1)}
= \frac{\depthz(\node_{2k})}{\pol(\node_{2k})}\,,
\end{align*}
so the state $s_k$ has a lower or equal cost below $\node_1$ than below $\node_2$.
Since this holds for any such $\act_{1:k}$, $\node_2$ can be safely cut, and by recurrence all cuts preserve the best-first ordering and lowest costs of states.
The rest of the proof is as in the non-Markovian case.
\end{proof}

\levints's cost function allows us to provide the following guarantee, which is an adaptation of Levin search's theorem~\citep{solomonoff1984optimum} to tree search problems.

\begin{theorem}\label{thm:levints}
Let $\targetset$ be a set of target nodes, then \levints{} with a policy $\pol$ ensures that the number of node expansions $\nodexp(\levints, \targetset)$ before reaching any of the target nodes is bounded by
\begin{align*}
\nodexp(\levints, \targetset) \leq \min_{\node\in\targetset} \frac{\depthz(\node)}{\pol(\node)}\,.
\end{align*}
\end{theorem}
\begin{proof}
From \cref{thm:bestfirstorder}, the first state of $\goalset$ to be expanded is the one of lowest cost, and with one of the nodes of lowest cost, that is, with cost
$c := \min_{\node\in\targetset} \depthz(\node)/\pol(\node)$.
Let $\tree_c$ be the current search tree when $\targetnode$ is being expanded.
Then all nodes in $\tree_c$ that have been expanded up to now have at most cost $c$.
Therefore at all leaves $\node\in\leaves(\tree_c)$ of the current search tree, $\depthz(\node)/\pol(\node) \leq c$.
Since each node is expanded at most once (each sequence of actions is tried at most once)
the number of nodes expanded by \levints{} until node $\targetnode$ is at most
\begin{align*}
\nodexp(\levints, \targetset) = 
|\nodeset(\tree_c)| \leq 
\sum_{\node\in\leaves(\tree_c)} \depthz(\node)
\leq \sum_{\node\in\leaves(\tree_c)} \pol(\node) c \leq c
= \min_{\node\in\targetset} \frac{\depthz(\node)}{\pol(\node)}
\end{align*}
where the first inequality is because each leaf of depth $\depthz$
has at most $\depthz$ ancestors,
the second inequality follows from $\depthz(n)/\pol(n)\leq c$,
and the last inequality is because $\sum_{\node\in\leaves(\tree_c)}\pol(\node) \leq 1$,
which follows from $\sum_{\node'\in\children(\node)}\pol(\node')=\pol(\node)$, that is,
each parent node splits its probability among its children, and the root has probability 1.
\end{proof}

The upper bound of \cref{thm:levints} is tight within a small factor for a tree like in
\cref{fig:chainandbin}, and is almost an equality when the tree splits at the root into multiple chains.




\section{Luby tree search: policy-guided unbounded sampling}


\paragraph{Multi-sampling}
When a good upper bound $\depthbound$ is known on the depth of a subset of the target nodes
with large cumulative probability, 
a simple idea is to sample trajectories according to $\pol$ (see \cref{lst:sampling})
of that maximum depth $\depthbound$ until a solution is found, if one exists.
Call this strategy \multits{} (see \cref{lst:multits}).
We can then provide the following straightforward guarantee.
\begin{theorem}\label{thm:multits}
The expected number of node expansions before reaching a node in $\targetset$ is bounded by
\begin{align*}
\expect[\nodexp(\multits(\infty, \depthbound), \targetset)] \leq \frac{\depthbound}{\pol^+_{\depthbound}}\ , \quad \quad\pol^+_{\depthbound} := \sum_{\substack{\node\in\targetset \\ \depthz(\node)\leq\depthbound}} \pol(\node)\,.
\end{align*}
\end{theorem}
\begin{proof}
Remembering that a tree search algorithm does not expand children of target nodes,
the result follows from observing that $\expect[\nodexp(\multits, \targetset)]$
is the expectation of a geometric distribution with success probability 
$\pol^+_{\depthbound}$ where each failed trial takes exactly $\depthbound$ node expansions and the success trial takes at most $\depthbound$ node expansions.
\end{proof}

This strategy can have an important advantage over \levints{} 
if 
there are many target nodes within depth bounded by $\depthbound$ with small individual probability but large cumulative probability.

The drawback is that if no target node has a depth shorter than the bound $\depthbound$, this strategy will never find a solution (the expectation is infinite), even if the target nodes have high probability according to the policy $\pol$.
Ensuring such target nodes can be always found leads to the \lubyts{} algorithm.

\paragraph{\lubyts}

Suppose we are given a randomized program $\prog$, that has an unknown distribution $\prob$ over the halting times (where halting means solving an underlying problem).
We want to define a strategy that can restart the program multiple times
and run it each time with a different allowed running time
so that it halts in as little cumulative time as possible in expectation.
\citet{luby1993speedup} prove that the optimal strategy is to run $\prog$ for running times of fixed lengths $\runt_\prob$ optimized for $\prob$; then either the program halts within $\runt_\prob$ steps, or it is forced to stop and is restarted for another $\runt_\prob$ steps and so on. 
This strategy has an expected running time of $\ell_\prob$, with $\frac{L_\prob}{4}\leq\ell_\prob\leq L_\prob = \min_{\runt\in\Naturals} \frac{\runt}{\cumu(\runt)}$ where $\cumu$ is the cumulative distribution function of $\prob$.
%
\citet{luby1993speedup} also devise a \emph{universal} restarting strategy based on a special sequence%
\footnote{\url{https://oeis.org/A182105}.}
of running times:
\begin{center}
1 1 2 1 1 2 4 1 1 2 1 1 2 4 8 1 1 2 1 1 2 4 1 1 2 1 1 2 4 8 16 1 1 2\ldots
\end{center}
They prove that the expected running time of this strategy is bounded by $192\ell_\prob (\log_2\ell_\prob + 5)$
and also prove a lower bound of $\tfrac{1}{8}\ell_\prob \log_2 \ell_\prob$ for any universal restarting strategy.
We propose to use instead the sequence%
\footnote{\url{https://oeis.org/A006519}. Gary Detlefs (ibid) notes that it can be computed with $\magicseq(n) := ((n\text{ XOR }n-1)+1)/2$ or with $\magicseq(n) := (n \text{ AND } -n)$ where $-n$ is $n$'s complement to 2.}
\magicseq:
\begin{center} 1 2 1 4 1 2 1 8 1 2 1 4 1 2 1 16 1 2 1 4 1 2 1 8 1 2 1 4 1 2 1 32 1 2\ldots\end{center}
which is simpler to compute and for which we can prove the following tighter upper bound.
\begin{theorem}\label{thm:luby}
For all distributions $\prob$ over halting times, 
the expected running time of the restarting strategy based on $\magicseq$ is bounded by $\min_\runt \runt + \frac{\runt}{\cumu(\runt)}\left(\log_2\frac{\runt}{\cumu(\runt)}+6.1\right)$, where $\cumu$ is the cumulative distribution of $\prob$.
\end{theorem}
The proof is provided in \cref{app:luby}.
We can easily import the strategy described above into the tree search setting (see \cref{lst:lubyts}), and provide the following result.

\begin{figure}
\begin{subfigure}[b]{0.4\textwidth}
\begin{lstlisting}[label=lst:multits,caption={Sampling of \texttt{nsims} trajectories of fixed depths $\depthbound \in\Naturals$.}]
def multiTS(nsims, $\depthbound$)
  for k := 1 to nsims
    if sample_traj($\depthbound$)
      return true
  return false
\end{lstlisting}
\end{subfigure}
~
\hspace*{0.9cm} 
\begin{subfigure}[b]{0.5\textwidth}
\begin{lstlisting}[label=lst:lubyts,caption={Sampling of \texttt{nsims} trajectories of depths that follow $\magicseq$, with optional coefficient $d_\text{min}\in\Naturals$.}]
def LubyTS(nsims, $d_\text{min}$=1)
  for k := 1 to nsims
    if sample_traj($d_\text{min}*\magicseq$(k))
      return true
  return false
\end{lstlisting}
\end{subfigure}
\end{figure}

\begin{theorem}\label{thm:lubyts}
Let $\targetset$ be the set of target nodes, then \lubyts($\infty$, 1)
with a policy $\pol$ ensures that the expected number of node expansions before reaching
a target node is bounded by
\begin{align*}
\expect[\nodexp(\lubyts(\infty, 1), \targetset)] \leq \min_{d\in\Naturals} d + \frac{d}{\pol^+_d}\left( \log_2 \frac{d}{\pol^+_d} + 6.1\right)\,,\quad\quad
&\pol^+_d := \sum_{\substack{\node\in\targetset\\\depthz(\node)\leq d}} \pol(\node)\,,
\end{align*}
where $\pol^+_d$ is the cumulative probability of the target nodes with depth at most $d$.
\end{theorem}
\begin{proof}
This is a straightforward application of \cref{thm:luby}:
The randomized program samples a sequence of actions from the policy $\pol$,
the running time $\runt$ becomes the depth $\depthz(\node)$ of a node $\node$,
the probability distribution $p$ over halting times becomes the probability of reaching a target node of depth $\runt$,
$\prob(\runt) = \sum_{\{\node\in\targetset, \depthz(\node)=\runt\}} \pol(\node)$,
and the cumulative distribution function $\cumu$ becomes $\pol^+_d$.
\end{proof}

Compared to \cref{thm:multits}, the cost of adapting to an unknown depth is 
an additional factor $\log(d/\pol^+_d)$.
The proof of \cref{thm:luby} suggests that the term $\log d$ is due to not knowing the lower bound on $d$, and the term $-\log \pol^+_d$ is due to not knowing the upper bound.
If a good lower bound $d_{\text{min}}$ on the average solution length is known, one can also multiply $\magicseq(n)$ by $d_{\text{min}}$ to avoid sampling too short trajectories as in \cref{lst:lubyts};
this may lessen the factor $\log d$ while still guaranteeing that a solution can be found if one of positive probability exists.
In particular, in the tree search domain, the sequence A6519 samples trajectories of depth 1 half of the time,
which is wasteful.
%
Conversely, in general it is not possible to cap $d$ at some upper bound, as this may prevent finding a solution as for $\multits$.
Hence the factor $-\log \pol^+_d$ remains, which is unfortunate since $\pol^+_d$ can easily be exponentially small with $d$.


\section{Strengths and weaknesses of \levints\ and \lubyts}\label{sec:compare}

Consider a ``needle-in-the-haystack problem'' represented by a perfect full and infinite binary search tree
where all nodes $\node$ have probability $\pol(\node)=2^{-\depth(\node)}$.
Suppose that the set $\targetset$ of target nodes contains a single node $\targetnode$
at some depth $d$. According to  
\Cref{thm:levints}, \levints{} needs to expand no more than $\depthz(\targetnode)2^{\depth(\targetnode)}$ nodes before expanding $\targetnode$. 
For this particular tree, the number of expansions is closer to $2^{\depth(\targetnode)+1}$ since there are only at most $2^{\depth(\targetnode)-1}$ nodes with cost lower or equal to $\cost(\targetnode)$.
\Cref{thm:lubyts} and the matching-order lower bound of \citep{luby1993speedup} suggest \lubyts{} may expand in expectation  $O(\depth(\targetnode)^2 2^{\depth(\targetnode)})$ nodes to reach $\targetnode$.
This additional factor of $\depth(\node)^2$ compared to \levints{} is a non-negligible price for needle-in-a-haystack searches.
For \multits, if the depth bound $\depthbound$ is larger than $\depthz(\targetnode)$,
then the expected search time is at most and close to $\depthbound 2^{\depth(\targetnode)}$,
which is a factor $\depth(\node)$ faster than \lubyts, unless $\depthbound \gg \depth(\targetnode)$.

Now suppose that the set of target nodes is composed of $2^{d-1}$ nodes, all at depth $d$.
Since all nodes at a given depth have the same probability,
\levints{} will expand at least $2^{d}$ and at most $2^{d+1}$ nodes before expanding any of the target nodes.
By contrast, because the cumulative probability of the target nodes at depth $d$ is
$1/2$, \lubyts{} finds a solution in $O(d\log d)$ node expansions, which is an exponential gain over \levints{}.
For \multits{} it would be $\depthbound$, which can be worse than $d\log d$ due to the need for  a large enough $\depthbound$.

\levints{} can perform state cuts 
if the policy is Markovian, which can substantially reduce the algorithm's search effort.  
For example, suppose that in the binary tree above every left child represents the same state as the root and thus is cut off from the search tree, leaving in effect only $2d$ nodes for any depth $d$.
If the target set contains only one node at some depth $d$,
even when following a uniform policy, \levints{} expands only those $2d$ nodes. By contrast,  \lubyts{} expands in expectation more than $O(2^d)$ nodes.
\levints{} has a memory requirement that grows linearly with the number of nodes expanded,
as well as a log factor in the computation time due to the need to maintain a priority queue to sort the nodes by cost. 
By contrast, \lubyts{} and \multits{} have a memory requirement that grows linearly with the solution depth, as they only need to store in memory the trajectory sampled. 
\levints's memory cost could be alleviated with an iterative deepening \citep{korf1985depth} variant with transposition table~\citep{ReinefeldM94}.


\section{Mixing policies and avoiding zero probabilities}

For both \levints{} and \lubyts{}, if the provided policy $\pol$ incorrectly assigns a probability too close to 0 to some sequences of actions, then
the algorithm may never find the solution.
To mitigate such outcomes, it is possible to `mix' the policy with the uniform policy so that the former behaves slightly more like the latter.
There are several ways to achieve this, each with their own pros and cons.

\paragraph{Bayes mixing of policies}
If $\pol_1$ and $\pol_2$ are two policies, we can build their Bayes average $\pol_{12}$ with prior $\alpha\in[0, 1]$ and $1-\alpha$ such that for all sequence of actions $\act_{1:t}$, $\pol_{12}(\act_{1:t}) = \alpha \pol_1(\act_{1:t}) + (1-\alpha)\pol_2(\act_{1:t})$.
The conditional probability of the next action is given by
\begin{align*}
\pol_{12}(\act_t | \act_{<t}) &= w_1(\act_{<t})\pol_1(\act_t | \act_{<t}) + 
w_2(\act_{<t})\pol_2(\act_t | \act_{<t}) \\
\text{with } w_1(\act_{<t}) &= 1-w_2(\act_{<t}) = \frac{\alpha \pol_1(\act_{<t})}{\alpha \pol_1(\act_{<t}) + (1-\alpha) \pol_2(\act_{<t})} = \alpha\frac{ \pol_1(\act_{<t})}{\pol_{12}(\act_{<t})}\,,
\end{align*}
where $w_1(\act_{<t})$ is the `posterior weight' of the policy $\pol_1$ in $\pol_{12}$.
This ensures that for all nodes $\node, \pol_{12}(\node) \geq \alpha \pol_1(\node)$ 
and $\pol_{12}(\node)\geq(1-\alpha)\pol_2(\node)$ which leads to the following refinement for \cref{thm:levints} for example (and similarly for \lubyts{}):
\begin{align*}
\nodexp(\levints, \targetset) \leq \min \left\{
\frac{1}{\alpha}
\min_{\node\in\targetset} \frac{\depthz(\node)}{\pol_1(\node)},
\frac{1}{1-\alpha}
\min_{\node\in\targetset} \frac{\depthz(\node)}{\pol_2(\node)}
\right\}\,.
\end{align*}
In particular, with $\alpha = 1/2$, \levints{} with $\pol_{12}$ is within a factor 2 of the best between \levints{} with $\pol_1$ and \levints{} with $\pol_2$.
More than two policies can be mixed together, leading for example to a factor $K$ compared to the best of $K$ policies when all prior weights are equal.
This is very much like running several instances of \levints{} in parallel, each with its own policy, except that (weighted) time sharing is done automatically.
For example, if the provided policy $\pol$ is likely to occasionally assign too low probabilities,
one can run \levints{} with a Bayes mixture of $\pol$ and the uniform policy, with a prior weight $\alpha$ closer to 1 if $\pol$ is likely to be far better than the uniform policy for most instances.

\newcommand{\epssteps}{\mathcal{K}}

\paragraph{Local mixing of policies, fixed rate}
Bayes mixing of two policies splits the search into 2 (mostly) independent searches.
But one may want to mix at a more `local' level: Along a trajectory $\act_{1:t}$, if the provided policy $\pol$ assigns high probability to almost all actions but a very low probability to a few ones, we may want to use a different policy just for these actions, and not for the whole trajectory.
Thus, given two policies $\pol_1$ and $\pol_2$ and $\eps\in[0, 1]$, the local-mixing policy $\pol_{12}$
is defined through its conditional probability $\pol_{12}(\act_t | \act_{<t}) := \eps \pol_1(\act_t | \act_{<t}) + (1-\eps)\pol_2(\act_t | \act_{<t})$.
Then for all $\act_{1:t}$,
\begin{align*}
\pol_{12}(\act_{1:t}) \geq
\underbrace{
\eps^{|\epssteps_1|}(1-\eps)^{t-|\epssteps_1|}}_{\text{penalty}}
\prod_{k\in \epssteps_1} \pol_1(\act_k|\act_{<k})
\prod_{k\notin \epssteps_1} \pol_2(\act_k|\act_{<k})\,,
\end{align*}
where $\epssteps_1$ is the set of steps $k$ where $\pol_1(\act_{k}|\act_{<k}) > \pol_2(\act_t |\act_{<k})$. 
This can be interpreted as `At each step $t$, $\pol$ must pay a factor of $\eps$ to use policy $\pol_1$ or a factor of $1-\eps$ to use $\pol_2$'.
This works well for example if $\eps\approx 0$ 
and $\epssteps_1$ is small, that is, the policy $\pol_2$ is used most of the time.
For example, $\pol_1$ can be the uniform policy, $\pol_1(\act_t|\act_{<t})\! =\! 1/|\actionset|$, and $\pol_2$ is a given policy that may sometimes be wrong.

\paragraph{Local mixing, varying rate}
The problem with the previous approach is that $\eps$ needs to be fixed in advance.
For a depth $d$, a penalty of the number of node expansions of $1/(1-\eps)^d\approx e^{\eps d}$ is large as soon as $d > 1/\eps$.
If no good bound on $d$ is known, one can use a more adaptive $1-\eps_d(\act_{1:t})=(t/(t+1))^\gamma$ with $\gamma \geq 0$: This gives $\prod_{k=1}^t (t/(t+1))^\gamma = 1/(t+1)^\gamma$, which means that the maximum price to pay to use only the policy $\pol_2$ for all the $t$ steps is at most $(t+1)^\gamma$, and the price to pay each step the policy $\pol_1$ is used is approximately $(t+1)/\gamma$.
The optimal value of $\eps$ can also be learned automatically using an algorithm such as Soft-Bayes~\citep{orseau2017softbayes} where the `experts' are the provided policies, but this may have a large probability overhead for this setup.

\section{Experiments: computer-generated Sokoban}

We test our algorithms on 1,000 computer-generated levels of Sokoban~\citep{racaniere2017imagination}
of 10x10 grid cells and 4 boxes.%
\footnote{The levels are available at \url{https://github.com/deepmind/boxoban-levels/unfiltered/test}.}
For the policy, we use a neural network pre-trained with A3C (details on the architecture and the learning procedure are in \cref{sec:network}).
We picked the best performing network out of 4 runs with different learning rates.
Once the network is trained, we compare the different algorithms using the same network's fixed Markovian policy.
Note that for each new level, the goal states (and thus target set) are different, whereas the policy does not change (but still depends on the state). 
We test the following algorithms and parameters: \lubyts(256,1), \lubyts(256,32), \lubyts(512, 32), \multits(1, 200), \multits(100, 200), \multits(200, 200), \levints{}. Excluding the small values (i.e., $\texttt{nsims} = 1$ and $d_{\text{min}}=1$), the parameters were chosen to obtain a total number of expansions within the same order of magnitude. 
In addition to the policy trained with A3C, we tested \levints, \lubyts, and \multits{} with a variant of the policy in which we add 1\% of noise to the probabilities output of the neural network. That is, these variants use the policy $\tilde{\pol}(\act|\node) = (1-\eps)\pol(\act|\node) + \eps\frac{1}{4}$ where $\pol$ is the network's policy and $\eps = 0.01$, to guide their search. These variants are marked with the symbol (*) in the table of results. 
We compare our policy tree search methods with a version of the LAMA  planner~\citep{Richter2010} that uses the lazy version of GBFS with preferred operators and queue alternation with the FF heuristic. This version of LAMA is implemented in Fast Downward~\citep{Helmert06thefast}, a domain-independent solver. We used this version of LAMA because it was shown to perform better than other state-of-the-art planners on Sokoban problems~\citep{XieNM12}. Moreover, similarly to our methods, LAMA searches for a solution of small depth rather than a solution of minimal depth.

\begin{figure}
    \centering
    \includegraphics[width=350px]{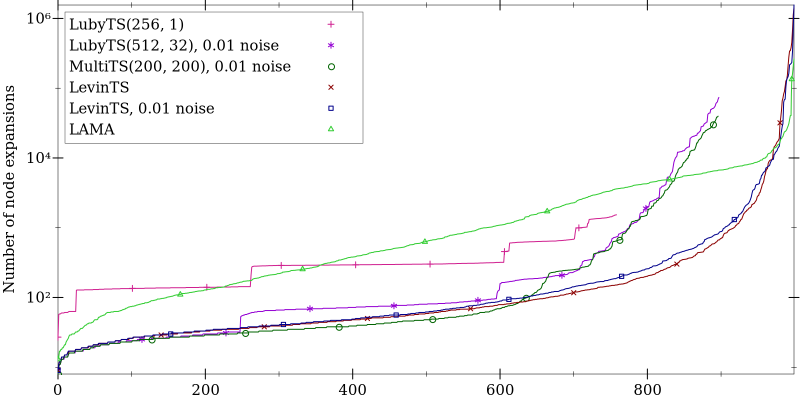}
    \caption{Node expansions for Sokoban on log-scale. The levels indices (x-axis) are sorted independently for each solver from the easiest to the hardest level. For clarity a typical run has been chosen for randomized solvers; see \cref{tab:results} for standard deviations.}
    \label{fig:expansions}
\end{figure}

\cref{tab:results} presents the number of levels solved (``Solved''), average solution length (``Avg. length''), longest solution length (``Max. length''), and total number of nodes expanded (``Total expansions''). The top part of the table shows the sampling-based randomized algorithms. In addition to the average values, we present the standard deviation of five independent runs of these algorithms. Since \levints{} and LAMA are deterministic, we present a single run of these approaches. \cref{fig:expansions} shows the number of nodes expanded per level by each method when the levels are independently sorted for each approach from the easiest to the hardest Sokoban level in terms of node expansions.  
The Uniform searcher (\levints{} with a uniform policy) with maximum 100,000 node expansions per level---and still with state cuts---can  solve no more than 9\% of the levels, which shows that the problem is not trivial.

\begin{table}
    \centering
    \caption{Comparison of different solvers on the 1000 computer-generated levels of Sokoban.
    For randomized solvers (shown at the top part of the table), the results are aggregated over 5 random seeds ($\pm$ indicates standard deviation). (*) Uses $\tilde{\pol}$ with $\eps=0.01$.}
    \label{tab:results}
\begin{tabular}{l r@{\,}l r@{\,}l r@{\,}l r@{\,}l}
\toprule
\textbf{Algorithm} & \multicolumn{2}{c}{\textbf{Solved}} & \multicolumn{2}{c}{\textbf{Avg. length}} & \multicolumn{2}{c}{\textbf{Max. length}} & \multicolumn{2}{c}{\textbf{Total expansions}} \\
\midrule
Uniform & 88 &  & 19 &  & 59 &  & 94,423,278 & \\
\midrule
LubyTS(256, 1) & 753 &$\pm$ 5 & 41.0 &$\pm$ 0.6 & 228 &$\pm$ 18.6 & 63,8481 &$\pm$ 2,434\\
LubyTS(256, 32) & 870 &$\pm$ 2 & 48.4 &$\pm$ 0.9 & 1,638.4 &$\pm$ 540.7 & 6,246,293 &$\pm$ 73,382\\
LubyTS(512, 32) & 884 &$\pm$ 4 & 54.8 &$\pm$ 4.2 & 3,266.6 &$\pm$ 1,287.8 & 11,515,937 &$\pm$ 211,524\\
LubyTS(512, 32) (*) & 896 &$\pm$ 2 & 50.7 &$\pm$ 2.5 & 1,975.6 &$\pm$ 904.5 & 10,730,753 &$\pm$ 164,410\\
MultiTS(1, 200) & 669 &$\pm$ 5 & 41.3 &$\pm$ 0.6 & 196.4 &$\pm$ 2.2 & 93,768 &$\pm$ 925\\
MultiTS(100, 200) & 866 &$\pm$ 4 & 47.8 &$\pm$ 0.5 & 199.4 &$\pm$ 0.5 & 3260536 &$\pm$ 57185\\
MultiTS(200, 200) & 881 &$\pm$ 1 & 47.9 &$\pm$ 0.7 & 196.4 &$\pm$ 2.3 & 5,768,680 &$\pm$ 116,152\\
MultiTS(200, 200) (*) & 895 &$\pm$ 3 & 48.8 &$\pm$ 0.4 & 198.8 &$\pm$ 1 & 5,389,534 &$\pm$ 45,085\\
\midrule
LevinTS & 1,000 && 39.8 &  & 106\phantom{.0} &  & 6,602,666 & \\
LevinTS (*) & 1,000 &  & 39.5 &  & 106\phantom{.0} &  & 5,026,200 & \\
LAMA & 1,000 &  & 51.6 &  & 185\phantom{.0} &  & 3,151,325 & \\
\bottomrule
\end{tabular}
\end{table}

For most of the levels, \levints{} (with the A3C policy) expands many fewer nodes than LAMA, but has to expand many more nodes on the last few levels. On 998 instances, the cumulative number of expansions taken by \levints{} is \textasciitilde{}2.7e6 nodes while LAMA expands \textasciitilde{}3.1e6 nodes. These numbers contrast with the number of expansions required by \levints{} (6.6e6) and LAMA (3.15e6) to solve all 1,000 levels. 
The addition of noise to the policy reduces the number of nodes expanded by \levints{} while solving harder instances at the cost of increasing the number of nodes expanded for easier problems (see the lines of the two versions of \levints{} crossing at the right-hand side of \cref{fig:expansions}). Overall, noise reduces from 6.6e6 to 5e6 the total number of nodes \levints{} expands (see \cref{tab:results}). 
\levints{} has to expand a large number of nodes for a small number of levels likely due to the training procedure used to derive the policy. That is, the policy is learned only from the 65\% easiest levels solved after sampling single trajectories---harder levels are never solved during training. Nevertheless, \levints{} can still solve harder instances by compensating the lack of policy guidance with search. 

The sampling-based methods have a hard time reaching 90\% success, but still
improves by more than 20\% over sampling a single trajectory. 
\lubyts(256, 32) improves substantially over \lubyts(256, 1) since many solutions have length around 30 steps. \lubyts(256, 32) is as good as \multits(200, 100) that uses a hand-tuned upper bound on the length of the solutions.
        
The solutions found by \levints{} are noticeably shorter (in terms of number of moves) than those found by LAMA. It is remarkable that \levints{} can find shorter solutions and expand fewer nodes than LAMA for most of the levels. This is likely due to the combination of good search guidance through the policy for most of the problems and \levints's systematic search procedure. By contrast, due to its sampling-based approach, \lubyts{} tends to find very long solutions. 

\citet{racaniere2017imagination} report different neural-network based solvers applied to a long sequence of Sokoban levels generated by the same system used in our experiments (although we use a different random seed to generate the levels, we believe they are of the same complexity).
\citeauthor{racaniere2017imagination}'s primary goal was not to produce an efficient solver per se, but to demonstrate how an integrated neural-based learning and planning system can be robust to model errors and more efficient than an MCTS baseline.
Their MCTS approach solves 87\% of the levels within approximately 30e6 node expansions (25,000 per level for 870 levels, and 500 simulations of 120 steps for the remaining 130 levels).
Although \levints\ had much stronger results in our experiments, we note that \citeauthor{racaniere2017imagination}'s implementation of MCTS commits to an action every 500 node expansions. By contrast, in our experimental setup, we assume that \levints\ solves the problem before committing to an action. This difference makes the results not directly comparable.
\citeauthor{racaniere2017imagination}'s second solver (I2A) is a hybrid model-free and model-based planning using a LSTM-based recurrent neural network 
with more learning components than our approaches. 
I2A reaches 95\% success within an estimated total of 5.3e6 node expansions (4,000 on average over 950 levels, and 30,000 steps for the remaining 50 unsolved levels; this counts the internal planning steps). 
For comparison, \levints{} with 1\% noise solves all the levels within the same total time (999 for \levints{} without noise). Moreover, \levints\  solves 95\% of the levels within a total of less than 160,000 steps, which is approximately 168 node expansions on average for solved levels, compared to the reported 4,000 for I2A. Moreover, it is also not clear how long it would take I2A to solve the remaining 5\%.



\section{Conclusions and future works}

We introduced two novel tree search algorithms for single-agent problems that are guided by a policy: \levints{} 
and \lubyts{}.
Both algorithms have guarantees on the number of nodes that they expand before reaching a solution (strictly for \levints{}, in expectation for \lubyts{}). 
\levints{} and \lubyts{} depart from the traditional heuristic approach to tree search by employing a policy instead of a heuristic function to guide search while still offering important guarantees.


The results on the computer-generated Sokoban problems using a pre-trained neural network show that these algorithms can largely improve through tree search upon the score of the network during training. Our results also showed that \levints\ is able to solve most of the levels used in our experiment while expanding many fewer nodes than a state-of-the-art heuristic search planner. In addition, \levints\ was able to find considerably shorter solutions than the planner. 

The policy can be learned by various means or it can even be handcrafted. In this paper we used reinforcement learning to learn the policy. 
However, the bounds offered by the algorithms could also serve directly as metrics to be optimized while learning a policy; this is a research direction we are interested in investigating in future works. 


\newpage
\paragraph{Acknowledgements}
 The authors wish to thank Peter Sunehag, Andras Gyorgy, R\'emi Munos, Joel Veness, Arthur Guez, Marc Lanctot, Andr\'e Grahl Pereira, and Michael Bowling for helpful discussions pertaining this research. Financial support for this research was in part provided by the Natural Sciences and Engineering Research Council of Canada (NSERC).

\bibliographystyle{abbrvnat}
\bibliography{biblio}

\newpage
\ifdefined\suppmat \setcounter{page}{2}\fi 
\begin{appendices}
\section{Network architecture and learning protocol}\label[appendix]{sec:network}

\begin{figure}
    \centering
    \includegraphics[width=200px,trim={0 0 0 25px},clip]{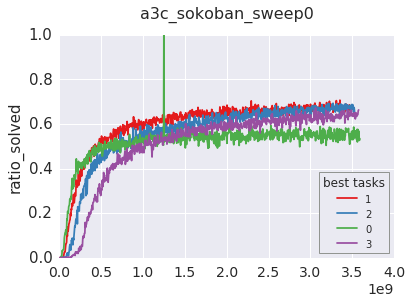}
    \caption{Learning curves of A3C for the 4 chosen learning rates (4e-4, 2e-4, 1e-4, 5e-5) on the Sokoban level generator.}
    \label{fig:network_learning}
\end{figure}

The network takes as input a 10x10x4 grid where the last dimension is for a binary encoding of the different attributes (wall, man, goal, box), which is passed through 2 convolutional layers ($4\times 4$ with $64$ channels, followed by $3\times 3$ with 64 channels as well), followed by a fully connected layer of 512 ReLU units.
The output layer provides logits for the 4 actions (up, down, left, right).
Training is performed using A3C~\citep{mni2016asynchronous} with a reward function giving a reward of -0.1 per step, +1 per box on a goal and -1 for the converse action, and +10 for solving the level (all boxes on goals),
with a discount factor of 0.99;
the optimizer used is RMSProp~\citep{Tieleman2012rmsprop} (no momentum, epsilon 0.1, decay 0.99),
with entropy regularization of 0.005.
During training, at each episode, the learner performs a single trajectory of length 100 (like \multits(1, 100)), receives the corresponding rewards, then moves on to the next episode. A single level is (very likely) never seen twice during training.
Similarly, it is very unlikely that a level of the 1000 test levels was seen during training.
We take the best performing network, which solves around 65\% of the levels
when sampling a single sequence of actions.
The network is trained for 3.5e9 steps (node expansions), which can seem to be a lot, however notice that
this is equivalent to fully searching a \emph{single} level of Sokoban (without state cuts) uniformly with 4 actions up to depth 16 (given that solutions are usually of depth more than 30).
The learning process was repeated for 4 learning rates (4e-4, 2e-4, 1e-4, 5e-5) (see \cref{fig:network_learning}).

\section{Another universal restarting strategy for Las Vegas programs}\label[appendix]{app:luby}

We use the sequence\footnote{\url{https://oeis.org/A006519}.}
of runtimes
 $f(n) := \text{A6519}(n)$:
\begin{center} 1 2 1 4 1 2 1 8 1 2 1 4 1 2 1 16 1 2 1\ldots\end{center}
\begin{align*}
\text{For all } n\in\Naturals:
f(n) := \begin{cases}
1&\text{ if } n \text{ is odd}, \\
2f(n/2) &\text{ o.w.}
\end{cases}
\end{align*}
It has the `fractal' property that $f(k2^n)= 2^nf(k)$ (since $f(k2^n)=2f(k2^{n-1})=\ldots=2^nf(k2^0)$),
for $k\in\Naturals$ and $n\in\Nonnegints$,
and it follows that $f(2^n) = 2^n$
and  $f(k2^n) \geq 2^n$.

At iteration $n$, the Las Vegas program is run for $f(n)$ steps.
For all $\runt>0$, if $f(n) \geq \runt$, then it has a probability at least $\cumu(\runt)$ of halting,
otherwise it does not halt and is forcibly stopped after $f(n)$ computations steps.
Let $\uprunt := 2^{\lceil \log_2 \runt\rceil}$ be the smallest power of 2 greater than or equal to $\runt$.
Then \cref{lem:A6519fractalshift} below tells us that for $c < \uprunt$ we have that
$f(k\uprunt + c) = f(c) \leq \uprunt/2 < \runt$, that is, between two consecutive factors of $\uprunt$,
$f(n) <\runt$.

Let $p_\text{halt}(n)$ denote the probability that the algorithm
halts exactly at the $n$th run,
and take $1\leq c  < \uprunt$ and $k \geq 0$,
then the expected number of computation steps $L$
(sum of the lengths of the runs)
 before halting is given by:
\begin{align*}
L_\text{univ}(\prob) &:= \sum_{n=1}^\infty
\left[
 \runt\, p_\text{halt}(n) + (1-p_\text{halt}(n)) f(n)\right]
 \underbrace{\prod_{j=1}^{n-1}(1-p_{\text{halt}}(j))}_{\substack{\text{probability of}\\\text{not halting before run $n$}}}
 \,.
\end{align*}
where $p_\text{halt}(n) = 0$ when $f(n) < \runt$, and $p_\text{halt}(n) = \cumu(\runt)$ otherwise.

We restate \cref{thm:luby} more precisely:
\begin{theorem}\label{thm:luby2}
For all distributions $\prob$ over halting times, 
the expected runtime of the universal restarting strategy based on $A6519$ is bounded by
\begin{align*}
L_\text{univ}(\prob) \leq 
\min_\runt \runt + \frac{\runt}{\cumu(\runt)}\left(\log_2\frac{\runt}{\cumu(\runt)}+6.1\right)\,,
\end{align*}
where $\cumu$ is the cumulative distribution of $\prob$.
\end{theorem}

\begin{proof}[Proof of \cref{thm:luby,thm:luby2}]
At step $n$, if $k$ is the number of past runs where $f(m) \geq \uprunt$ (with $m < n$),
then $\prod_{j=1}^{n-1}(1-p_{\text{halt}}(j)) = (1-\cumu(\runt))^k$
then with $1\leq c  < \uprunt$ and $\gamma := 1-\cumu(\runt)$:
\begin{align*}
L_\text{univ}(\prob)&=\sum_{n=0}^\infty \begin{cases}
\gamma^k f(n)&\text{if } n=k\uprunt + c  \quad(\text{\ie} f(n) < \runt)\\
\gamma^k p\runt + \gamma^{k+1} f(n)&\text{if }n=k\uprunt+\uprunt\,,\\
\end{cases} \\
&= \sum_{n=0}^\infty \begin{cases}
\gamma^k f(c)&\text{if } n=k\uprunt + c\\
\gamma^k p \runt + \gamma^{k+1}\uprunt f(k+1)&\text{if }n=k\uprunt+\uprunt\,.
\end{cases}
\end{align*}
where we used $f((k+1)\uprunt)=\uprunt f(k+1)$ (remembering that $\uprunt$ is a power of 2) and \cref{lem:A6519fractalshift} for $f(k\uprunt+c) = f(c)$.
Since $f(n)=f(c) < \runt$ when $n=k\uprunt+c$, we can decompose $L_\text{univ}(\prob)$ into the steps
where $f(n) < \runt$ and the rest:
\begin{align*}
L_\text{univ}(\prob) &= L^{<} + L^\geq \\
L^< &:= \sum_{k=0}^\infty \gamma^k\sum_{c=1}^{\uprunt-1} f(c)=\frac{1}{1-\gamma}\sum_{c=1}^{\uprunt-1} f(c) = \frac{\uprunt}{2\cumu(\runt)}\log_2 \uprunt\quad \text{(\cref{lem:A6519sum})}\\
L^\geq &:= \sum_{k=0}^\infty \gamma^k(1-\gamma)\runt + \gamma^{k+1}\uprunt f(k+1)
= \runt + \uprunt\sum_{k=1}^\infty \gamma^k f(k) \\
&\leq \runt + \frac{\uprunt}{\cumu(\uprunt)}\left(
\frac{1}{e} + \frac{1}{\ln 2} + \frac{1}{2}\log_2 \ln 16 + \frac{1}{2}\log_2 \frac{1}{\cumu(\uprunt)}\right)
\end{align*}
where we used \cref{lem:A6519generative} on the last line with $\gamma = 1-\cumu(\runt)$.
Finally, since $\uprunt =2^{\lceil\log_2 \runt\rceil}\leq 2\runt$
and $\cumu(\uprunt) \geq \cumu(\runt)$
and $\lceil\log_2 \runt\rceil \leq \log_2 \runt + 1$: 
\begin{align*}
L &\leq \runt + \frac{\runt}{\cumu(\runt)}\left(\log_2 \runt + 1 + \frac{2}{e} + \frac{2}{\ln 2} + \log_2 \ln 16 + \log_2 \frac{1}{\cumu(\runt)}
\right)\\
 &\leq \runt + \frac{\runt}{\cumu(\runt)}\left(\log_2 \frac{\runt}{\cumu(\runt)} + 6.1\right)
\end{align*}
which proves the result.
\end{proof}

\begin{lemma}\label{lem:A6519fractalshift}
For $f = $A6519,
with $k\in\Nonnegints, n\in\Nonnegints, a\in\Naturals, b\in\Nonnegints$
and $a2^b < 2^n$, and with $a$ odd, then
\begin{align*}
f(k2^n + a2^b) = f(a2^b) = 2^b\,.
\end{align*}
\end{lemma}
\begin{proof}
Since $a$ is odd, then so is $k2^{n-b} + a$, and so
$f(k2^n + a2^b) = f(2^b(k2^{n-b} + a)) = 2^b f(k2^{n-b} + a) = 2^b$.
\end{proof}
Hence, for all numbers between two adjacent factors of $2^n$,
$f(k2^n + c) =f(c) \leq 2^{n-1}$.

\begin{lemma}\label{lem:A6519sum}
For $n \in\Naturals$ and $f=$A6519,
\begin{align*}
\sum_{c=1}^{2^n-1} f(c) = n2^{n-1}.
\end{align*}
\end{lemma}
\begin{proof}
If $n \geq 1$ and using \cref{lem:A6519fractalshift} again at $2^{n-1}$:
\begin{align*}
\sum_{c=1}^{2^n-1} f(c) &= \sum_{c=1}^{2^{n-1}-1} f(c) + f(2^{n-1}) + \sum_{c=2^{n-1}+1}^{2^{n}-1} f(c) \\
&= 2^{n-1} + 2\sum_{c=1}^{2^{n-1}-1} f(c)\\
&=\ldots=2^02^{n-1} + 2^12^{n-2} + 2^22^{n-3}+ \ldots +2^{n-1}2^0 + 2^n\sum_{c=1}^{2^{0}-1} f(c)\\
&=n2^{n-1}\,.
\end{align*}
\end{proof}

\begin{lemma}\label{lem:A6519expt2n}
Let $f=$A6519,
then for $k\in\Naturals, n\in\Nonnegints, c\in\Nonnegints$:
\begin{align*}
f(k) = 2^n\quad \Leftrightarrow\quad k=(2c+1)2^n\,.
\end{align*}
\end{lemma}
\begin{proof}
Since any number $k$ can be uniquely written in the form
$k=(2c+1)2^a$, and $f((2c+1)2^a)=2^af(2c+1)=2^a$ with $a\in\Nonnegints$,
then $f(k)=2^n \Leftrightarrow a=n$.
\end{proof}

\begin{lemma}\label{lem:2ng2n}
For $\gamma \in [0, 1)$, 
\begin{align*}
\sum_{n=0}^\infty 2^n \gamma^{2^n} \leq \frac{1}{\ln \frac{1}{\gamma}}\left(\frac{1}{e} + \frac{\gamma}{\ln 2} \right)\,.
\end{align*}
\end{lemma}
\begin{proof}
Let $h(x) := 2^x \gamma^{2^x}$ for $x\in\Reals$,
then $h'(x) = \ln(2)2^{x}\gamma^{2^{x}}(2^{x}\ln\gamma + 1)$ where
$h'(x_0) = 0$ for the unique $x_0$ such that $2^{x_0} = \frac{1}{\ln\frac{1}{\gamma}}$
and since $\ln \gamma < 0$, we have that $h'(x)$ is positive for $x < x_0$ and negative for $x > x_0$.
Thus $h$ is unimodal, and since furthermore $h(x)$ is positive the sum can be upper bounded by the integral of the continuous function plus its maximum:
\begin{align*}
\sum_{n=0}^\infty h(n) &\leq \int_{0}^\infty h(x) \text{d}x + \max_x h(x)\,,\\
\max_x h(x) &= h(x_0) = \frac{1}{\ln \frac{1}{\gamma}}\frac{1}{e}\,,\\
\int_{0}^\infty 2^x \gamma^{2^x} \text{d}x & =\frac{1}{\ln 2} \int_0^\infty 2^x\ln 2 \gamma^{2^x} \text{d}x
= \frac{1}{\ln 2}\int_{1}^\infty \gamma^y \text{d}y = \frac{\gamma}{\ln2\ln\frac{1}{\gamma}}\,,
\end{align*}
where we used integration by substitution.
Adding the two terms finishes the proof.
\end{proof}

\begin{lemma}\label{lem:sumg2n}
For $\gamma \in[0, 1)$ and $a\geq 1$: 
\begin{align*}
\sum_{n= 0}^\infty \gamma^{2^n} \quad\leq
\gamma\left\lceil\log_2 \frac{1}{\log_2 \frac{1}{\gamma}}\right\rceil + 1
\quad\leq \log_2 \frac{1}{\ln \frac{1}{\gamma}} + \log_2 \ln 16
\,.
\end{align*}
\end{lemma}
\begin{proof}
Let $N =\min \left\{n\in\Nonnegints: \gamma^{2^N} \leq \half \right\}=\left\lceil\log_2 \frac{1}{\log_2 \frac{1}{\gamma}}\right\rceil$, then
\begin{align*}
\sum_{n= 0}^\infty \gamma^{2^n} &=
\sum_{n= 0}^{N-1} \gamma^{2^n} + \sum_{n= N}^\infty \gamma^{2^n} \\
&\leq N\gamma + \sum_{n=0}^\infty \left(\gamma^{2^N}\right)^{2^n}
\leq N\gamma + \sum_{n=0}^\infty 2^{-2^n}
 \leq N\gamma + 1\\
&\leq
\left\lceil\log_2 \frac{1}{\log_2 \frac{1}{\gamma}}\right\rceil + 1 \\
&\leq
\log_2 \frac{1}{\log_2 \frac{1}{\gamma}} + 2\,.
\end{align*}
Extracting $\log_2 \ln 2$ finishes the proof.
\end{proof}

\begin{lemma}\label{lem:A6519generative}
Let $f=$A6519 and $\gamma \in[0, 1)$. 
Then
\begin{align*}
\sum_{k=1}^\infty \gamma^k f(k)
\leq \frac{1}{1-\gamma}\left(
\frac{1}{e} + \frac{1}{\ln 2} + \frac{1}{2}\log_2 \ln 16 + \frac{1}{2}\log_2 \frac{1}{1-\gamma}\right)
\,.
\end{align*}
\end{lemma}
\begin{proof}
Since $f(n)$ is a power of 2 for all $n\in \Naturals$, we regroup the runs
by powers of 2:
\begin{align*}
\sum_{k=1}^\infty \gamma^k f(k)
&=\sum_{n=0}^\infty 2^n\sum_{k=1}^\infty \gamma^k\indicator{f(k)=2^n} \\
&=\sum_{n=0}^\infty 2^n\sum_{c=0}^\infty \gamma^{(2c+1)2^n} \quad \text{(\cref{lem:A6519expt2n})} \\
&=\sum_{n=0}^\infty 2^n\gamma^{2^n}\sum_{c=0}^\infty \left(\gamma^{2^{n+1}}\right)^c
=\sum_{n=0}^\infty 2^n\gamma^{2^n}\frac{1}{1-\gamma^{2^{n+1}}} \\
&\leq \sum_{n=0}^\infty 2^n \gamma^{2^n}\left(1+\frac{\gamma}{2^{n+1}(1-\gamma)}\right) \quad\text{(\cref{lem:exptolin})}\\
&=\sum_{n=0}^\infty 2^n \gamma^{2^n} +\frac{1}{2}\frac{\gamma}{1-\gamma}\sum_{n=0}^\infty \gamma^{2^n}\\
&\leq \frac{1}{1-\gamma}\left(\frac{1}{e}+ \frac{\gamma}{\ln 2} +
\frac{\gamma}{2}
+\frac{\gamma^2}{2}\left(\log_2\ln 4 + \log_2 \frac{1}{1-\gamma}\right)\right) \\
&\leq \frac{1}{1-\gamma}\left(
\frac{1}{e} + \frac{1}{\ln 2} + \frac{1}{2}\log_2 \ln 16 + \frac{1}{2}\log_2 \frac{1}{1-\gamma}\right)
\end{align*}
where we used \cref{lem:2ng2n} and \cref{lem:sumg2n}
 on the second to last line together with $\ln\frac{1}{\gamma} \geq 1-\gamma$.
\end{proof}

\begin{lemma}\label{lem:exptolin}
For $\gamma \in[0, 1)$ and $a\geq 1$: 
\begin{align*}
\frac{1}{1-\gamma^a} \leq 1+\frac{1}{a}\frac{\gamma}{1-\gamma}\,.
\end{align*}
\end{lemma}
\begin{proof}
For $\epsilon > 0$ and $a\geq1$, it can be shown that $(1+\eps)^a \geq 1+a\eps$.
Then, taking $\gamma := \frac{1}{1+\eps}$:
\begin{align*}
(1+\eps)^a \geq 1+a\eps &\quad\Leftrightarrow\quad (1+\eps)^a - 1 \geq a((1+\eps)-1) \\
&\quad\Leftrightarrow\quad
\frac{1}{(1+\eps)^a -1} \leq \frac{1}{a((1+\eps)-1)} \\
&\quad\Leftrightarrow\quad
\frac{1}{\gamma^{-a} -1} \leq \frac{1}{a(\gamma^{-1}-1)} \\
&\quad\Leftrightarrow\quad
\frac{\gamma^a}{1-\gamma^{a}} \leq \frac{\gamma}{a(1-\gamma)} \\
&\quad\Leftrightarrow\quad
\frac{1}{1-\gamma^a} \leq 1+\frac{1}{a}\frac{\gamma}{1-\gamma}\,,
\end{align*}
which proves the result.
\end{proof}

\end{appendices}

\end{document}